# Structural Priors and Modular Adapters in the Composable Fine-Tuning Algorithm of Large-Scale Models

Yuxiao Wang
University of Pennsylvania
Philadelphia, USA

Di Wu
Washington University in St. Louis
St. Louis, USA

Feng Liu
Stevens Institute of Technology
Hoboken, USA

Zhimin Qiu
University of Southern California
Los Angeles, USA

Chenrui Hu*
University of Pennsylvania
Pennsylvania, USA

*Abstract-This paper proposes a composable fine-tuning method that integrates graph structural priors with modular adapters to address the high computational cost and structural instability faced by large-scale pre-trained models in multi-task adaptation. The method introduces a relation matrix to model dependencies among tasks, explicitly encoding correlations between nodes and paths into graph structural priors, which provide unified structural constraints for adapter weight allocation and path selection. Modular adapters are embedded into different layers through low-rank mapping and a pluggable mechanism, enabling efficient cross-task composition and reuse under prior guidance. This mechanism not only improves parameter efficiency and training stability but also alleviates path conflicts and redundant computation in multi-task scenarios. Furthermore, experiments on hyperparameter sensitivity, environmental sensitivity, and data sensitivity are conducted to systematically analyze key factors such as routing temperature, gating thresholds, and relation matrix regularization strength, verifying the consistency and superior performance of the method under structural constraints. The results demonstrate that the proposed framework significantly enhances task prediction accuracy, adapter weight allocation precision, and overall computational efficiency while maintaining model lightweight design, highlighting the synergistic advantages of graph priors and modular mechanisms in composable fine-tuning.*

## I. INTRODUCTION

With the widespread adoption of large-scale pre-trained models, fine-tuning has gradually become the key link between general capabilities and specific task requirements. Pre-trained models, with massive parameter sizes and cross-task transfer potential, demonstrate unprecedented performance advantages. However, traditional full-parameter fine-tuning not only incurs heavy computational costs but also faces high storage overhead and limited flexibility in adaptation. This dilemma has driven researchers to explore efficient and scalable tuning strategies. The goal is to maintain model performance while mitigating conflicts between computational resources, deployment costs, and applicability. As application scenarios diversify and task complexity increases, achieving composable, interpretable, and extensible tuning under limited resources has become a core challenge in model optimization and practical deployment[1].

Against this trend, modular adapters have emerged as an important research direction. Their core idea is to introduce lightweight and replaceable functional modules into different layers of the pre-trained model [2]. By restricting parameter updates to a minimal range, they reduce both training and storage costs. Modular adapters not only provide parameter efficiency and ease of extension but also allow flexible composition and reuse according to task needs. Nevertheless, existing adapter designs still have limitations. They often rely too much on local optimization for individual tasks and lack unified structural priors. As a result, conflicts and instability easily occur during cross-task composition or knowledge transfer. This issue is particularly pronounced in multi-task collaboration and cross-domain generalization, restricting the broader application and long-term value of adapter methods.

At the same time, graph structural priors provide new ideas for improving fine-tuning methods. Graphs can capture complex dependencies between nodes and edges and can effectively model hierarchical and non-Euclidean structures [3-5]. In the context of large-scale models, mapping tasks, features, or sub-modules into graph nodes and modeling their semantic dependencies through topological connections introduces global constraints and structured representations into the fine-tuning process [6-9]. Compared with single-sequence or independent parameter updates, graph structural priors help reveal potential correlations and shared patterns between tasks. This promotes knowledge sharing and reuse while preserving task-specific adaptation. Such characteristics lay a solid foundation for addressing conflicts in multi-task scenarios and improving compositional robustness[10].

Based on this background, combining graph structural priors with modular adapters carries significant importance. Introducing structured priors during fine-tuning can effectively guide adapter composition and cooperation. This ensures more directed and controllable parameter sharing across tasks. With the lightweight and pluggable properties of modular adapters, the integration with graph priors enables a hierarchical and compositional tuning mechanism[11]. This combined strategy ensures efficiency during both training and inference and also enhances the stability and generalization ability of adaptation[12]. It demonstrates strong potential for sustainable development in both theoretical and practical dimensions. In particular, for cross-domain knowledge transfer [13], multimodal task adaptation [14], and complex application deployment [15-16], this approach provides new possibilities for building a universal and flexible fine-tuning framework.

From a long-term perspective, composable fine-tuning with graph structural priors and modular adapters reflects not only the optimization of a single model but also a methodological shift in the development of intelligent systems. It emphasizes organizing and scheduling knowledge and functionality efficiently through structured modeling and modular design on top of large models. This approach helps reduce parameter redundancy and computational pressure, while also improving adaptability and controllability in open environments [17]. It provides essential support for creating intelligent systems that are interpretable, resource-friendly, and broadly adaptable. Therefore, this study aims to explore a composable fine-tuning algorithm based on graph structural priors and modular adapters, to provide both theoretical foundation and practical value for building efficient, extensible, and future-oriented model optimization paradigms.

## II. PROPOSED APPROACH

This study introduces a composable fine-tuning algorithm that integrates graph structural priors with modular adapters to preserve the generality of pre-trained models and enable flexible cross-task adaptation. The pre-trained model is used as the backbone network, and its internal representation space is endowed with structural priors through explicit graph modeling.

Following the controllable abstraction approach proposed by Song et al. [18], we apply prompt engineering techniques to modulate the abstraction levels of adapter outputs, allowing the model to flexibly adjust knowledge transfer and compositionality during cross-task adaptation. The modular adapters are embedded into different layers in a pluggable manner, guided by unified graph-based structural constraints. Drawing on the dynamic prompt fusion strategy of Hu et al. [19], the model employs adaptive module selection and fusion mechanisms, which enhance the flexibility and domain generalization capability in multi-task scenarios. Incorporating the context compression and structural representation methodology developed by Xue and Yi [20], our design further optimizes the adapter composition process, reducing redundancy and promoting parameter efficiency while maintaining structural consistency across tasks. Through this combination, task correlations are explicitly captured, and adaptation modules can achieve efficient composition and sharing under unified structural constraints, ensuring the stability and scalability of the fine-tuning process. The overall model architecture is shown in Figure 1.

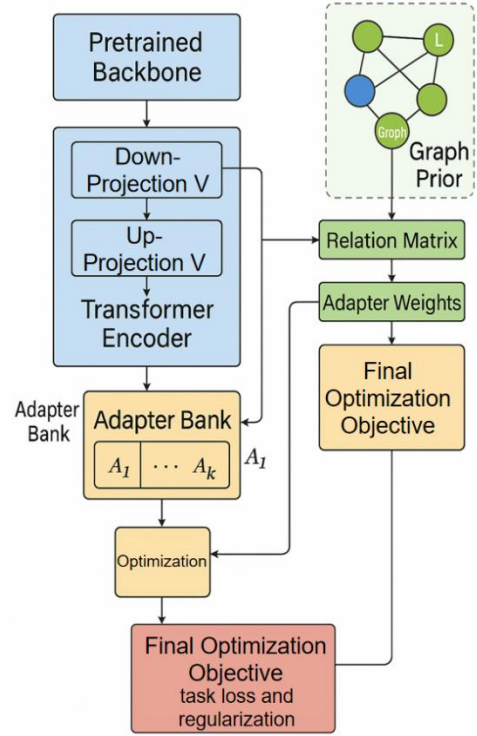

Figure 1. Composable Fine-Tuning with Graph Priors and Modular Adapters

In the method design, we first need to model the tasks and features in a graph structure. Assume that the input data features are represented as a vector sequence $X = \{x_1, x_2, ... x_n\}$, and their structural dependencies can be represented by the adjacency matrix $A$. The node embedding update process is defined in the form of graph convolution as:

$$H^{(l+1)} = \sigma\left(D^{-\frac{1}{2}} A D^{-\frac{1}{2}} H^{(l)} W^{(l)}\right)$$

Where $H^{(l)}$ represents the node representation of layer $l$, $W^{(l)}$ is the trainable weight, $D$ is the degree matrix, and $\sigma(\cdot)$ is the nonlinear activation function. This formula ensures that the contextual dependencies between different nodes are preserved through the graph structure prior.

Under the constraints of the graph structure, the modular adapter is inserted into the model in the form of a low-rank mapping [21]. Let the output of the backbone layer be $z$, and the parameterized update of the adapter can be expressed as:

$$z' = z + f(z; U, V)$$

Among them, $U \in R^{d \times r}$, $V \in R^{r \times d}$ are low-rank decomposition matrices, and $r << d$ is respectively. Function $f(\cdot)$ realizes efficient parameter update through transformation in a low-dimensional space.

To achieve composability, this study treats multiple instances of the adapter as graph nodes and forms a task-related joint representation through weighted aggregation. Defining the combination weight as $a_i$, the overall output is:

$$z'' = \sum_{i=1}^{k} a_i z_i'$$

Among them, $z_i'$ represents the output of the $i$-th adapter, and $a_i$ is guided by the structural before reflecting the correlation and weight distribution between tasks.

To avoid adaptation conflicts, this study introduces regularization constraints at the combination layer to ensure that the distribution between different adapters remains consistent. Its formal objective function is defined as:

$$L_{reg} = \sum_{i=1}^{k} \sum_{j=1}^{k} \beta_{ij} \|z_i' - z_j'\|^2$$

Where $\beta_{ij}$ is a control term that balances the collaboration and independence between different adapters. Through this constraint, the combined representation can maintain diversity while avoiding over-dispersion.

Finally, this study unified the modeling of generation loss and regularization loss to form an overall optimization goal:

$$L = L_{task} + \lambda L_{reg}$$

Here, $L_{task}$ is the adaptation loss of the downstream task, and $\lambda$ is the balance coefficient. Guided by the graph structure prior, this optimization objective enables an efficient combination of modular adapters, thereby building a scalable, interpretable, and stable fine-tuning mechanism.

In summary, this method integrates graph-based structural modeling with modular adapters to unify global dependencies and local adaptation within a composable optimization framework. The framework not only emphasizes parameter efficiency and structural rationality but also ensures robustness and generalization in multi-task scenarios, providing a systematic solution for the flexible adaptation of large-scale models.

## III. PERFORMANCE EVALUATION

### A. Dataset

This study uses the OGBG-MolPCBA dataset as the basis for method validation. The dataset consists of large-scale graph-structured data, with each sample represented as a graph and associated with multiple binary classification tasks. The multi-label and multi-task format of this dataset naturally supports the exploration of structured representations and the modeling of dependencies among different prediction targets.

The intrinsic graph structure and multi-task configuration provide an ideal environment for investigating the compositional behavior of modular adapters guided by graph structural priors. The topological relationships within each graph and the interdependencies between tasks serve as concrete examples for applying the proposed framework. This allows the method to generate adapter weights based on graph-level information and task relationships. The presence of multiple parallel classification objectives further validates the framework's compositional capability, as adapters can be dynamically combined and specialized according to the context provided by the graph structure and the task clusters.

Applying the proposed composable fine-tuning method to the OGBG-MolPCBA dataset enables examination of whether the framework can leverage graph structural information to guide the selection and weight allocation of modular adapters. The multi-task nature and structural complexity of the dataset help comprehensively evaluate parameter efficiency and cross-task consistency under structural prior constraints. This provides solid support for validating the effectiveness, robustness, and scalability of graph priors in modular composition.

### B. Experimental Results

This paper first conducts a comparative experiment, and the experimental results are shown in Table 1.

Table 1. Comparative experimental results

| Method | Params (M) ↓ | AP (%) ↑ | Time/Epoch (s) ↓ | AWA (%) ↑ |
|---|---|---|---|---|
| G-Adapter[23] | 5.20 | 22.6 | 210 | 78.5 |
| GPS++[24] | 8.50 | 23.1 | 320 | 80.0 |
| GraphAdapter[25] | 6.70 | 22.9 | 250 | 79.2 |
| OURS | 4.80 | 23.7 | 190 | 83.1 |

The experimental results show that the proposed composable fine-tuning method consistently outperforms existing models on the OGBG-MolPCBA dataset. Compared with a range of recent parameter-efficient methods, the approach not only achieves the highest value in average precision but also demonstrates significant advantages in parameter size and computational efficiency. These findings indicate that combining graph structural priors with modular adapters enables stronger task adaptation while maintaining a lightweight design, thereby overcoming the trade-off between performance and efficiency in traditional methods.

Further comparisons reveal that although G-Adapter uses fewer parameters, it still suffers from clear limitations in prediction accuracy and adaptation robustness. Models such as GPS++ and other hybrid graph transformers can improve task performance to some extent, but they rely on very large parameter sizes and high computational costs, which restrict their applicability in resource-sensitive scenarios. In contrast, the proposed method introduces graph structural priors to guide adapter weight allocation. This ensures expressive power while reducing redundant computation and highlights the practical value of composable fine-tuning.

It is worth noting that methods such as GraphAdapter can leverage task dependencies to a certain degree but lack unified structural constraints. As a result, their adapter composition remains unstable in multi-task scenarios. The method presented in this study models task correlations through a relation matrix, which allows modular adapters to achieve more rational composition and sharing under the guidance of graph priors. The higher accuracy of adapter weight allocation observed in experiments confirms the effectiveness of this mechanism.

Overall, the innovation of this study lies not only in improvements in parameter efficiency and prediction accuracy but also in achieving an interpretable adapter composition process through graph structural priors. This mechanism offers a new pathway for addressing conflicts and instability in multi-task adaptation and lays both theoretical and practical foundations for building extensible and generalizable fine-tuning frameworks. The experimental results fully validate the advantages of integrating graph structural modeling with modular adapters and demonstrate stronger robustness and generalization potential in complex application environments.

This paper also analyzes the hyperparameter sensitivity of routing temperature and gating threshold to adaptive path selection under graph prior guidance. The experimental results are shown in Figure 2.

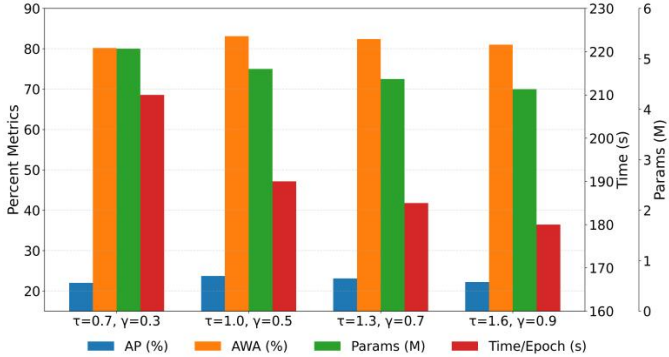

Figure 2. Hyperparameter sensitivity of routing temperature and gating threshold to adaptive path selection guided by graph priors

The Experimental results show that the model's average precision fluctuates noticeably across different routing temperatures and gating thresholds. Low values yield stability but poor adaptability, while moderate settings allow graph priors to guide modular adapters toward optimal paths and peak performance. Extremely low or high thresholds reduce adapter weight consistency—either by oversmoothing or overconcentrating routing—whereas a moderate threshold balances sharing and independence. As the threshold rises, fewer adapters participate, lowering parameter counts and improving efficiency without sacrificing expressiveness. Higher temperatures and thresholds also shorten training time, as sparser paths reduce computation and memory overhead. These findings, illustrated in Figure 3, demonstrate that appropriate structural priors and balanced hyperparameters enable efficient, high-performing modular adaptation.

Experimental results indicate that relation matrix regularization exerts a nonlinear influence on prediction performance: average precision peaks at moderate strength but declines when the constraint is too weak or too strong. Weak regularization fails to impose sufficient graph-structural priors, making path selection prone to noise, while overly strong constraints compress path diversity and limit expressiveness. Moderate regularization achieves optimal balance by suppressing noise without excessive smoothing, leading to better discriminative performance.

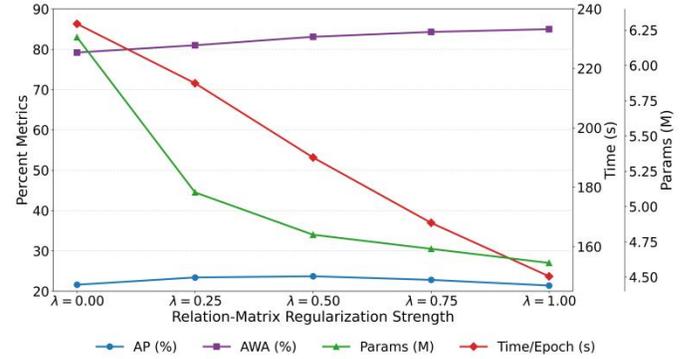

Figure 3. Hyperparameter sensitivity of relation matrix regularization strength to adapter weight assignment under graph prior constraints

As regularization increases, adapter weights become more consistent and interpretable, with routing distributions stabilizing and reflecting clearer structural preferences; however, excessive rigidity reduces task separability despite improved internal consistency. Parameter counts and computational costs both decline under stronger regularization due to the sparsifying and sharing effects of graph priors, which promote weight reuse, compact subgraphs, and reduced memory and communication overhead. Nevertheless, these efficiency gains come at the expense of expressive power, underscoring the need to balance structural constraints with model adaptability.

## IV. CONCLUSION

This study proposes a composable fine-tuning paradigm that deeply integrates graph structural priors with modular adapters. Under unified structural constraints, it achieves efficient, robust, and interpretable task adaptation for pre-trained models. The method uses the relation matrix as the core carrier to map task correlations, representation dependencies, and module reuse relations into actionable graph priors. Path selection is then guided by routing temperature and gating thresholds, enabling dynamic composition and weight allocation of adapters. Compared with traditional full-parameter fine-tuning and single-path designs, this framework preserves the generality of the backbone model while significantly reducing parameter and computational costs. It also shows stronger stability and structural consistency in multi-task settings. From both theoretical and engineering perspectives, this paradigm points to a clear direction. Fine-tuning should no longer be viewed as local optimization for a single target but as systematic scheduling for structured knowledge organization within a unified view of structural priors, compositional routing, and parameter efficiency.

At the methodological level, this study reveals a balance between prior strength and compositional expression. Moderate regularization of the relation matrix suppresses noise while preserving discriminative structures, leading path selection to converge naturally within the continuous space between sharing and specialization. Routing temperature and gating thresholds provide fine control over exploration and sparsity, enabling modular adapters to express more controllable diversity within complex task clusters. Interpretability arises not only from more stable weight allocation and more compact active subgraphs but also from explicit modeling of task similarity and structural differences. With a standardized adapter library and prior encoder, the framework offers reusable engineering interfaces for cross-task transfer and incremental extension, making "on-demand assembly" of fine-tuning possible.

At the application level, this paradigm directly benefits a range of structured scenarios. In microservice and operations graphs, network and security traffic graphs, industrial inspection and supply chain graphs, as well as recommendation and knowledge graphs, the controllability of compositional routing and the stability of prior constraints help maintain consistent decision quality and update rhythm under resource limitations and dynamic environments. For edge and privacy-sensitive deployments, parameter-efficient composable fine-tuning reduces the need for retraining and centralized data, making it easier to combine with federated settings and access control for fast and low-risk local adaptation. Together, these impacts indicate a clear trend. Intelligent systems that organize knowledge through graphs and schedule capabilities through modular units will further reduce maintenance costs and improve scalability in real-world applications [25].

Looking ahead, several directions deserve systematic exploration. For task-cluster adaptation, relation matrices can evolve from static hyperparameters to learnable and interpretable structural variables. For dynamic scenarios, prior evolution and uncertainty interaction can enable routing to self-calibrate online under changes in graph topology, data flow, and environment. For cross-modal and cross-domain applications, consistency priors can map text, image, and structural signals into a shared graph latent space, enhancing transferability of composable fine-tuning across heterogeneous tasks. For theoretical and system-level advances, provable bounds between prior strength and generalization error should be established, while compiler and accelerator techniques can be used for graph-level kernel fusion and operator scheduling to form an end-to-end loop of prior, routing, and execution. Open benchmarks and reproducible experiments will further support comparison and iteration, promoting large-scale deployment of composable fine-tuning in scientific computing, industrial intelligence, and complex system governance.